\documentclass[10pt,twocolumn,letterpaper]{article}

\usepackage{wacv}              

\usepackage[accsupp]{axessibility} 
\usepackage{graphicx}
\usepackage{amsmath}
\usepackage{amssymb}
\usepackage{booktabs}
\usepackage{multirow}
\usepackage[inline]{enumitem}
\usepackage{array}
\newcolumntype{M}[1]{>{\centering\arraybackslash}m{#1}}

\usepackage[pagebackref,breaklinks,colorlinks]{hyperref}

\usepackage[capitalize]{cleveref}
\crefname{section}{Sec.}{Secs.}
\Crefname{section}{Section}{Sections}
\Crefname{table}{Table}{Tables}
\crefname{table}{Tab.}{Tabs.}

\def\wacvPaperID{2365} 
\def\confName{WACV}
\def\confYear{2025}

\begin{document}
\title{ConvMixFormer- A Resource-efficient Convolution Mixer for Transformer-based Dynamic Hand Gesture Recognition}

\author{Mallika Garg, Debashis Ghosh,  and Pyari Mohan Pradhan \\
  Indian Institute of Technology, Roorkee,  India-247667\\
  \tt\small(mallika@ec, debashis.ghosh@ece, pmpradhan@ece).iitr.ac.in }
\maketitle

\begin{abstract}
Transformer models have demonstrated remarkable success in many domains such as natural language processing (NLP) and computer vision. With the growing interest in transformer-based architectures, they are now utilized for gesture recognition. So, we also explore and devise a novel ConvMixFormer architecture for dynamic hand gestures. The transformers use quadratic scaling of the attention features with the sequential data, due to which these models are computationally complex and heavy. We have considered this drawback of the transformer and designed a resource-efficient model that replaces the self-attention in the transformer with the simple convolutional layer-based token mixer. The computational cost and the parameters used for the convolution-based mixer are comparatively less than the quadratic self-attention. Convolution-mixer helps the model capture the local spatial features that self-attention struggles to capture due to their sequential processing nature. Further, an efficient gate mechanism is employed instead of a conventional feed-forward network in the transformer to help the model control the flow of features within different stages of the proposed model. This design uses fewer learnable parameters which is nearly half the vanilla transformer that helps in fast and efficient training. The proposed method is evaluated on NVidia Dynamic Hand Gesture and Briareo datasets and our model has achieved state-of-the-art results on single and multimodal inputs. We have also shown the parameter efficiency of the proposed ConvMixFormer model compared to other methods. The source code is available at https://github.com/mallikagarg/ConvMixFormer.
\end{abstract}

\section{Introduction}
Deep learning models, particularly Convolutional Neural Networks (CNNs), have demonstrated the ability to automatically learn hierarchical representations of raw input data, through a series of convolutional and pooling layers~\cite{s22062405}. With time, attention-based mechanisms came into existence that focus more on important features. Earlier, deep  based~\cite{garg2021multiview, mallika2022two}, RNN-based and LSTM-based~\cite{mittal2019modified} models were used for recognizing continuous gestures.  Later, transformer network~\cite{vaswani2017attention} was used for capturing information from sequential data~\cite{d2020transformer} which takes single and multimodal inputs using a late fusion approach. Although transformers are very efficient in various applications, they pose some limitations in terms of performance and model complexity, specifically when used for computer vision tasks. For an image as input data, the sequential length corresponds to the number of tokens or patches extracted from the image. Even for a   224 $\times$ 224  image size, the number of tokens can be significant, resulting in a quadratic increase in attention computations, since the attention scales quadratically in the self-attention module of the transformer model.

Therefore, while designing the proposed model, we consider the transformer model's drawbacks for gesture recognition. We propose a novel token mixer that involves convolution for mixing spatial tokens of the input gesture image called ConvMixFormer, which generalizes with fewer parameters than the vanilla transformer. The main idea of using convolution as a token mixer as it is efficient in terms of parameters and FLOPs. Meanwhile, convolution also helps extract local features from the input, enabling it to capture fine-grained details and spatial relationships. This is required in gesture recognition tasks as hand and finger movements are involved in gestures, which can be of different sizes depending on the signer's age. By applying convolutions across the sequence, the model can capture interactions between adjacent tokens, allowing it to capture spatial dependencies and patterns. Convolutional layers also exhibit translation equivalence, meaning that if the input is translated, the output will also be translated by the same amount, which makes the model rotation and translation invariant.

Overall, using convolution as a token mixer in transformer models offers several advantages, including local feature extraction, complexity reduction, interaction between tokens, and comparable performance with fewer parameters. However, in the literature, ConvMixer~\cite{trockman2023patches} is also proposed which uses depthwise convolution to mix spatial locations followed by pointwise convolution to mix channel locations. While depthwise convolutions offer advantages such as reduced computational cost and model size, they also pose some disadvantages compared to standard convolutions: 
\begin{enumerate*}[label=(\alph*)]
\item  Depthwise separable convolutions consider independent input channels for processing which does not allow it to capture complex spatial patterns as in the case of convolutions. Therefore, it may struggle to capture inter-channel dependencies effectively. This limitation could result in a reduced ability to capture cross-channel spatial information.
\item Depthwise separable convolutions involve separate operations for each channel, leading to more memory accesses compared to standard convolutions, where each filter interacts with all input channels simultaneously.
\end{enumerate*}

Keeping in view the limitations of the transformer model and mixers proposed in the literature, we proposed the standard convolution as the token mixer for dynamic gesture recognition. We also propose the use of a Gated mechanism in the feed-forward network named a Gated Depthwise Feed Forward Network, (GDFN) to control the flow of information within different stages of the proposed transformer model. 

Thus, the key contributions are:
\begin{enumerate}
\item We design a lightweight resource-efficient convolution-based transformer model, ConvMixFormer, for dynamic hand gesture recognition.  
\item We propose a novel convolution token mixer to efficiently replace the attention mechanism in the transformer with the convolution layer that enables the model to capture complex spatial patterns from the input. 		
\item  We propose the use of a Gated Depthwise Feed Forward Network (GDFN) which helps the control of the information flow in a sequential transformer model. This helps to focus on relevant features while suppressing irrelevant information, potentially improving the model's robustness and generalization.
\item ConvMixFormer proves its efficacy on the Briareo and NVGesture dataset by achieving state-of-the-art performance with significantly fewer parameters on single and multimodal inputs.
\end{enumerate}

\section{Related Work}

\begin{figure*}[tb]				
\centerline{\includegraphics[scale=.65]{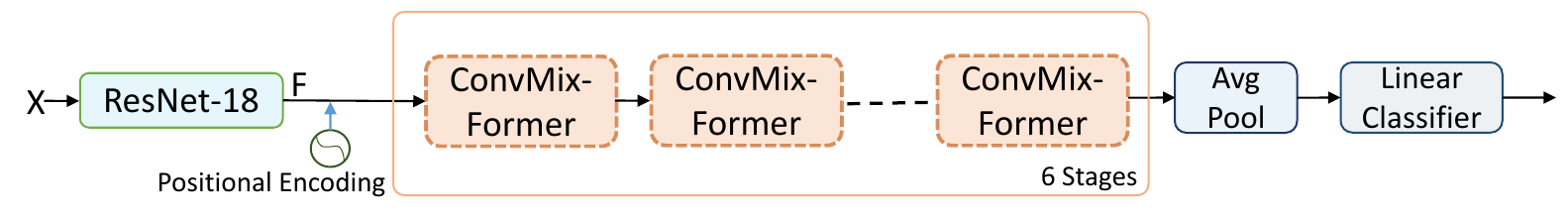}}
\caption{The overview of the proposed pipeline for dynamic gesture recognition. The input $X$ can be any input eg. normal, color, depth, optical flow, or infrared images. Our pipeline uses 6 ConvMixFormer stages. }
\label{fig3}
\end{figure*}

\subsection{Token Mixer}
To efficiently replace high computation self-attention in the transformer, MetaFormer~\cite{yu2022metaformer} proposed a generalized structure that acts as a token mixer. The extremely weak token mixer which uses a non-parametric pooling operator for token mixing, named PoolFormer achieves high performance, with a very simple network. Since then, the researchers have paid attention to designing an efficient token mixer to replace the vanilla self-attention-based token mixer, which is computationally expensive. ConvMixer~\cite{trockman2023patches} uses only standard convolutions (Depthwise convolution followed by pointwise convolution) to achieve the mixing steps in the transformer model by simply operating on the patches, similar to the vision transformer. This helps to capture the distant features with large receptive fields same as ViT. Similarly, multilayer perceptron (MLP) is also explored for token mixing in~\cite{tolstikhin2021mlp}. Each token represents a spatial location or a specific element in the input data, and the token-mixing MLPs operate independently on each channel (feature dimension) of the data. By processing individual columns of the data table independently, these MLPs allow for communication and interaction between different spatial locations or tokens. Motivated by MLP-mixer, a novel spatial-shift MLP ($S^2$-MLP)~\cite{yu2022s2} has been designed which only contains channel-mixing MLP. Separable Vision Transformer (SepViT)~\cite{li2022sepvit} uses 2 modules for token mixing, Depthwise Self-Attention for each pixel token and   Pointwise Self-Attention to fuse information from different channels. 

Though the exploration of Token Mixers was initially in the spatial domain, it was later explored in the frequency domain. FNet~\cite{lee2022fnet} uses linear transformations (Fourier transform) that replace self-attention with a parameter-free Fourier transform for mixing tokens. Wavemix~\cite{jeevan2024wavemixsr} uses wavelet transform,  a multi-level 2D discrete wavelet transform, and convolution for mixing in the frequency domain which can generalize which fewer parameters.  Similarly, Multiscale Wavelet Pooling Transformer (GestFormer)~\cite{garg2024gestformer} utilizes PoolFormer as the baseline with input first transformed using wavelet transform which is used for gesture recognition. PoseFormerV2~\cite{zhao2023poseformerv2} uses Discrete Cosine Transform (DCT) to narrow the gap between time and frequency domain features and utilize the DCT coefficients to encode multiple levels of temporal information for the input sequential data. The frequency representations help to remove high-frequency components which are nothing but noise components.

\subsection{Convolution in Transformers}
Integrating convolutional neural networks (CNNs) with transformers has become a popular approach to address the computational complexity and leverage the strengths of both architectures. Some other methods uses convolutional layers with the transformer models e.g.~POTTER~\cite{zheng2023potter},  Convolutional Vision Transformer (CvT)~\cite{wu2021cvt}, Swin Transformer\cite{liu2021swin}, CSwin~\cite{dong2022cswin}, CeiT~\cite{yuan2021incorporating}, Unifying  CNNs~\cite{li2023uniformer}, CoFormer~\cite{deshmukh2024textual}, etc.

Convolutions with transformers help the model to take advantage of convolution like scale, shift, dimension reduction with depth in layers, etc along with the merits of transformers like global context, attention mechanism, etc~\cite{wu2021cvt}.   While Convolution neural networks (CNNs) can reduce local redundancy,  they cannot learn global features due to small receptive fields. Unifying  CNNs with transformers can leverage the benefits of both CNNs and transformers which can capture global and local features~\cite{li2023uniformer}.  Incorporating CNNs into transformers not only helps in less computations and learning local-global features but also helps in faster convergence~\cite{xiao2021early,yuan2021incorporating}.  LeVit~\cite{graham2021levit} introduces attention bias with convolution to integrate position information in the transformer for faster convergence. LeViT uses convolutional embedding instead of the patch-wise projection used in ViT.

With convolution, the spatial dimension decreases and the depth increases, which is beneficial for transformer architecture when convolution is used before attention block~\cite{yuan2021hrformer}.   Swin Transformer\cite{liu2021swin} were introduced with shifted windows using convolutions in transformers. The original self-attention block overlaps single low-resolution features at each stage of the transformer. Swin transformer creates a hierarchy of scaled non-overlapping windows which makes the model flexible to various scales. Axial attention is another attention mechanism that aligns multiple dimensions of the attention into the encoding and the decoding settings~\cite{ho2019axial}. Later, another hierarchical design that uses cross attention in horizontal and vertical strips to form the cross-shaped window was introduced in~\cite{dong2022cswin}. CSwin introduces Locally-enhanced Positional Encoding (LePE), which is a position encoding scheme that handles local positions. This provides CSwin with a strong power to model input into attention features with less computation.

\subsection{Transformer for Gesture Recognition}
Hand gesture recognition has captured researchers attention as it can be used in many applications like Human-computer interaction, Virtual and augmented reality~\cite{ohkawa2023assemblyhands}, Sign language recognition~\cite{kumar2017coupled},  Autonomous vehicles~\cite{garg2021deep}. Transformers, initially designed for NLP tasks like translation, have been successfully applied to gesture recognition tasks.  Later, transformers were used in gesture recognition~\cite{de-coster-etal-2020-sign}. In~\cite{d2020transformer}, multimodal input sequences can be processed for dynamic gesture recognition, which uses Video transformers~\cite{neimark2021video} as the base model. The attention in the spatial dimension is not enough to model the video sequence in dynamic hand gestures, so local and global multi-scale attention was proposed both locally and globally~\cite{s22062405}. The local attention extracts the information of the hand and the global attention learns the human-posture context. A combination of convolutions with self-attention was proposed in~\cite{s22062405}  for fusing spatial and temporal features for multimodal dynamic gesture recognition.  Recurrent 3D convolutional neural networks are also used in conjugation with transformer models for end-to-end learning for egocentric gesture recognition~\cite{Cao_2017_ICCV}. In~\cite{vaswani2017attention}, since the transformer must know the ordering of the sequence input, sinusoidal position embedding is added with the input. Initially, methods that used a transformer for gesture recognition used sinusoidal positional encoding but later, a new  Gated Recurrent Unit (GRU)-based positioning scheme was incorporated into the Transformer networks~\cite{aloysius2021incorporating}. Transformer-based models have also been explored for Dynamic Hand Gesture Recognition~\cite{garg2023multiscaled}. This model also uses single as well as multimodal inputs which are either raw or derived from the raw data.

\section{Method}
\subsection{Overview}\label{formats}
The proposed ConvMixFormer is designed for a dynamic hand gesture recognition task that takes a sequence of frames of a video hand gesture as input to predict what gesture is being performed. 
The input is fed to the ResNet-18~\cite{he2016deep} model to get the frame level features of the input gesture sequence denoted by $f_t^m$ from the $m^{th}$ frame.  For complete sequence, these frame level features are concatenated which can  be denoted as a function $F$ over complete input sequence $X$, such that,
\begin{equation}
\label{eqn:100}
F(X) = f_t^1 \oplus f_t^2 \oplus \dots \oplus   f_t^m 
\end{equation}
where $F: \mathbb{R}^{m \times w \times h \times c} \rightarrow \mathbb{R}^{m \times k}$ is a function of feature extractor with each feature of size $k$.

A pretrained ResNet-18 on ImageNet database~\cite{deng2009imagenet} is used for feature extraction. The output features from ResNet are of dimensions $N = B\times T \times D$, where $B$ is the batch size, $T$ is the number of frames and $D$ is the dimension of features extracted from ResNet-18 which is the same as the input of the first stage of the proposed ConvMixFormer model. We pass the input from all the convolutional layers of the ResNet-18 model and then through the average pool layer, the required features are obtained. These features are then fed to the proposed ConvMixFormer blocks as shown in Fig.~\ref{fig3}. The input layer of the model is made to adapt for all the input modalities as proposed in~\cite{molchanov2016online}. 

 The encoded information from the ConvMixFormer is then average pooled over all the frames as, 
\begin{equation}
\label{eqn:5}
H(x) = AvgPool(E(x)),
\end{equation}
where $E(x)$ is the complete transformer encoder output which comprises of 6 ConvMixFormer stages and $H(x)$ is the average pooling operation over the $m$ frames. The $H(x)$ output is then passed through a linear classifier to get the output probability distribution over $n$ classes. 

\subsection{ConvMixFormer}\label{40}
Our ConvMixFormer model uses a convolution layer instead of a self-attention mechanism conventionally used in traditional transformers~\cite{vaswani2017attention}. The self-attention is computed using a scaled-dot product which has quadratic complexity that increases with long sequences meaning the computational cost increases quadratically with the length of the input sequence. While the self-attention mechanism is powerful for capturing long-range dependencies, its quadratic complexity can be a bottleneck, particularly for vision-based tasks involving very long sequences. To handle this drawback, we have proposed a simple yet efficient pure convolution-based transformer model, ConvMixFormer.

\begin{figure}[tb]				
\centerline{\includegraphics[scale=.48]{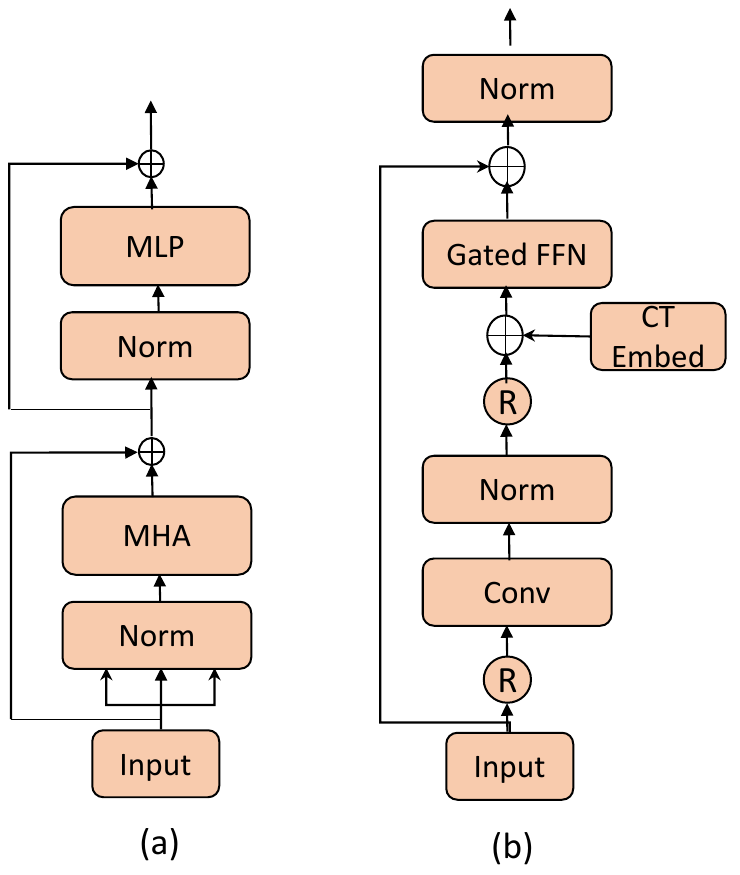}}
\caption{(a). The conventional Transformer model consists of 2 main blocks- multi-head attention (MHA) and Feedforward network (MLP). (b). The proposed ConvMixFormer model replaces the attention mechanism with a convolution layer to mix the spatial tokens with lesser parameters and uses a gate mechanism to selectively filter the information to pass to the next stage of the transformer model. Here,  \textcircled{R} denotes the reshaping of dimensions and CT Embed represents the class token embedding.}
\label{fig2}
\end{figure}

The motivation while designing this model involves less computation cost with comparable performance. Thus, we replace the self-attention operation with a simple convolution operation involving a convolution layer followed by a batch normalization layer as shown in Fig~\ref{fig2}. Convolution helps capture the fine-grained spatial details and intricate patterns present in the input data.  It enables interactions between neighboring tokens within the sequence. By applying convolutions across the sequence, the model can capture interactions between adjacent tokens.  

The features obtained from the ResNet, $F(X)$ are given as input to the proposed convolution token mixer of the proposed model. Class Token embedding followed by WaveMixSR~\cite{jeevan2024wavemixsr} is also added to the output from the proposed token mixer. After the token mixing block, the features are fed to the  Gated Depthwise Feed Forward Network (GDFN) block, described in Section~\ref{4}. Then, we add a skip connection to the output of the GDFN and normalize the complete output which can be written as, 
\begin{equation}
\label{eqn:4}
E(x) = Norm(x + GDFN(Conv(x))),
\end{equation}
where $x$ is the input feature to the ConvMixFormer model.

\subsection{Gated Dconv Block}\label{4}
Introducing a Gated Deconvolution Feedforward Network (GDFN) within a transformer block is an interesting approach to enhance the model's ability to filter and selectively process information. Gated mechanisms, popularized by architectures like LSTM  and GRU have shown effectiveness in controlling the flow of information in sequential models. Thus, we have also incorporated a gate mechanism in our model to focus on relevant and important features while suppressing noise or irrelevant information.  By controlling the flow of information, the model can learn more informative representations, leading to better feature extraction and higher-level representations, which may improve the performance on the model. We follow~\cite{zamir2022restormer}, to incorporate depth-wise convolutions within the Gated mechanism to effectively encode and process information.

Mathematically, the gating mechanism with depth-wise convolution can be represented as,
\begin{equation}
\label{eqn:3}	
Gating(\textbf{P}) =\phi(W^1_d W^1_p(\textbf{P})) \odot W^2_d W^2_p(\textbf{P}))
\end{equation}
here, $\textbf{P}$ is the input feature, $ \odot$ denotes the element-wise multiplication and $\phi $ represents the GELU activation. 

\begin{table*}[tb]
\caption{Results for different modalities on NVGesture~\cite{molchanov2016online}  and Briareo~\cite{manganaro2019hand} dataset. \# is the number of input modalities used. The best results for a given number of inputs are shown in bold for GateFormer.}
\centering
\begin{tabular}{c|ccccc|cc|cc}
\hline
\multirow{3}{*}{\#}&\multicolumn{5}{c|}{Input data}& \multicolumn{4}{c}{Accuracy} \\ \cline{2-10}
&\multirow{2}{*}{Color} &\multirow{2}{*}{ Depth}& \multirow{2}{*}{IR} & \multirow{2}{*}{Normals} & \multirow{2}{*}{Optical flow} & \multicolumn{2}{c|}{NVGesture} & \multicolumn{2}{c}{Briareo}\\  \cline{7-10}
&&&&&&Transformer~\cite{d2020transformer} &Ours & Transformer~\cite{d2020transformer} &Ours\\
\hline

\multirow{5}{*}{1} 
&\checkmark &&&&&    76.50\%&76.04\%& 90.60\%&\textbf{98.26}\%  \\
&&\checkmark &&&&    83.00\%&\textbf{80.83}\%& 92.40\%&97.22\%\\
&&&\checkmark&&&    64.70\%&63.54\%& 95.10\%&97.22\%\\
&&&&\checkmark&&    82.40\%&80.21\%& 95.80\%&97.57\%\\ 
&&&&&\checkmark&   72.00\%&74.17\%&-&96.88\%\\ \hline

\multirow{10}{*}{2}
&\checkmark&\checkmark&&&&  84.60\% &81.47\%&94.10\%&98.26\%\\
&\checkmark&&\checkmark&&&  79.00\%&77.29\%&95.50\%&\textbf{98.61}\%\\
&&\checkmark&\checkmark&&&  81.70\%&80.50\%&95.10\%&97.57\%\\
&\checkmark&&&\checkmark&&  84.60\%&80.29\%&96.50\%&98.26\%\\
&&\checkmark&&\checkmark&&  87.30\%&81.95\%&96.20\%&97.22\%\\
&&&\checkmark&\checkmark&&  83.60\%&77.18\%&97.20\%&97.97\%\\ 
&\checkmark&& &&\checkmark& -&77.46\%&-&97.92\%\\
&&\checkmark&&&\checkmark&  -&\textbf{82.16}\%&-&97.92\%\\ 
&&&\checkmark&& \checkmark& -&76.14\%&-&97.22\%\\
&&&&\checkmark&\checkmark& - &79.69\%&-&97.92\%\\\hline

\multirow{10}{*}{3} 
&\checkmark&\checkmark&\checkmark&&& 85.30\% &83.20\%&95.10\%&97.57\%\\
&\checkmark&\checkmark&&\checkmark&& 86.10\%&83.61\%&95.80\%&97.57\%\\
&\checkmark&&\checkmark&\checkmark&& 85.30\%&80.71\%&96.90\%&\textbf{98.64}\%\\
&&\checkmark&\checkmark&\checkmark&& 87.10\%&\textbf{84.02}\%&97.20\%&96.88\%\\ 

&\checkmark&\checkmark&&&\checkmark& -&83.82\%&-&97.22\%\\ 
&\checkmark&&\checkmark&&\checkmark& -&79.25\%&-&98.26\%\\ 
&\checkmark&&&\checkmark&\checkmark& -&81.12\%&-&97.92\%\\
&&\checkmark&\checkmark&&\checkmark& -&82.99\%&-&96.18\%\\ 
&&&\checkmark&\checkmark&\checkmark& -&82.16\%&-&97.57\%\\ 
&&\checkmark&&\checkmark&\checkmark& -&83.82\%&-&96.18\%\\ 
\hline

\multirow{5}{*}{4}
&\checkmark&\checkmark&\checkmark&\checkmark&&  87.60\%&85.27\%&96.20\%&97.57\%\\
&\checkmark&\checkmark&\checkmark&&\checkmark&  -&83.40\%&-&97.57\%\\
&\checkmark&\checkmark&&\checkmark&\checkmark&  -&84.23\%&-&97.57\%\\ 
&\checkmark&&\checkmark&\checkmark&\checkmark&  -&82.99\%&-&\textbf{98.26}\%\\ 
&&\checkmark&\checkmark&\checkmark&\checkmark&  -&\textbf{85.49}\%&-&97.57\%\\

\hline
5&\checkmark&\checkmark &\checkmark&\checkmark&\checkmark& -&85.48\%&-&97.57\%\\
\hline

\end{tabular}
\label{tab1}
\end{table*}

\subsection{Multi-Modal Late Fusion}\label{9}
In dynamic hand gesture recognition, the multi-model method processes the images of different modalities which include RGB, depth, and IR images, to enhance the accuracy and reliability of the model. The images of different modalities are given as input to the model to predict the probability of each class. The late fusion of the different modalities helps the model to capture the large and diverse aspect of the input data. The late fusion also helps to tackle the different challenges such as occlusion, viewpoints, and variation in the lighting condition. Following~\cite{d2020transformer} for late fusion, probability scores received from each modality is combined independently. We select the maximum probability obtained from individual model input to calculate the final prediction as,
\begin{equation}
\label{eqn:6}	
y = \arg\max_j \sum_{i}^n P(\omega_j|x_i),
\end{equation}
where $n$ represents the number of modalities over which the results are to be aggregated, and  $P(\omega_j|x_i)$ is the probability distribution of the  $i^{th}$ frames of a given input, which belongs to class $\omega_j$.

\section{Experiments and Discussion}\label{6}
\subsection{Datasets}\label{77}
\textbf{NVGesture:}
This dataset~\cite{molchanov2016online} is intended for touch-less driver controlling and it is developed for human-computer interaction through dynamic hand gestures. The dataset is acquired using multiple sensors and viewpoints to record depth, color, and IR images of the hand gestures. In the process of data collection, 20 subjects participated and captured 25 different types of gestures. Each subject used their left hand to control the steering wheel and right hand to perform gestures looking into the screen of the simulator. In total, 1532 samples were collected under a controlled environment with dim and bright lighting. Among all these, 1050 samples were used for training and 482 for testing purposes.

\begin{table}
\caption{Comparison results for single modality on NVGesture dataset~\cite{molchanov2016online} }
\centering
\begin{tabular}{ccc}
\hline
Input modality&Method& Accuracy \\ 
\hline
\multirow{11}{*}{Color} &Spat. st. CNN~\cite{simonyan2014two} &54.60\%  \\
&iDT-HOG~\cite{wang2016robust}& 59.10\%\\
&Res3ATN~\cite{dhingra2019res3atn}& 62.70\%\\
&C3D~\cite{tran2015learning}&69.30\%\\ 
&R3D-CNN~\cite{molchanov2016online}& 74.10\%\\
&GestFormer~\cite{garg2024gestformer}&75.41\%\\ 
&GPM~\cite{fan2021multi}&75.90\% \\
&PreRNN~\cite{yang2018making}&76.50\% \\
&Transformer~\cite{d2020transformer}& 76.50\%\\
&I3D~\cite{wang2016robust}&78.40\%\\
&\textbf{ConvMixFormer}& \textbf{76.04\%}\\
\hline

&Human~\cite{molchanov2016online}&88.40\% \\   \hline

\multirow{9}{*}{Depth}& SNV~\cite{yang2014super}& 70.70\%\\
&C3D~\cite{tran2015learning}&78.80\%\\ 
&GestFormer~\cite{garg2024gestformer}&80.21\%\\ 
&R3D-CNN~\cite{molchanov2016online}& 80.30\%\\
&I3D~\cite{wang2016robust}&82.30\%\\
&Transformer~\cite{d2020transformer}&83.00\%\\
&ResNeXt-101~\cite{kopuklu2019real}& 83.82\%\\
&PreRNN~\cite{yang2018making}&84.40\% \\
&MTUT~\cite{abavisani2019improving}&84.85\%\\
&\textbf{ConvMixFormer}& \textbf{80.83\%}\\

\hline	

\multirow{7}{*}{Optical flow}&iDT-HOF~\cite{vadisaction}& 61.80\% \\
&Temp. st. CNN~\cite{simonyan2014two} & 68.00\%\\
&Transformer~\cite{d2020transformer}& 72.00\%\\
&GestFormer~\cite{garg2024gestformer}&72.61\%\\ 
&iDT-MBH~\cite{vadisaction} & 76.80\%\\
&R3D-CNN~\cite{molchanov2016online} & 77.80\%\\
&I3D~\cite{wang2016robust} & 83.40\%\\
&\textbf{ConvMixFormer}&\textbf{74.17\%}\\
\hline

\multirow{3}{*}{Normals}&GestFormer~\cite{garg2024gestformer}&81.66\%\\ 
&Transformer~\cite{d2020transformer} &82.40\% \\
&\textbf{ConvMixFormer}& \textbf{80.21\%}\\ \hline

\multirow{3}{*}{Infrared}&R3D-CNN~\cite{molchanov2016online}& 63.50\%\\
&GestFormer~\cite{garg2024gestformer}&63.54\%\\ 
& Transformer~\cite{d2020transformer}&  64.70\% \\
&\textbf{ConvMixFormer}&\textbf{63.54\%}\\ \hline

\end{tabular}
\label{tab2}
\end{table}	

\textbf{Briareo:}
This dataset~\cite{manganaro2019hand} is collected in an automotive context for establishing human-computer interaction. To capture the dynamic hand gesture, the device is equipped with multiple sensors to capture depth, RGB, and motion images. RGB camera captures images with frame rates of 30fps, depth sensor which is a time-of-flight (ToF) sensor, captures with frame rates of 45fps, and stereo camera with 200fps. This dataset contains images from 12 different gesture classes which are captured by 40 subjects including 33 males and 7 females.

\subsection{Implementation Details}\label{7}
The proposed ConvMixFormer model is implemented using PyTorch version 1.7.1 on a system with an Nvidia GeForce GTX 1080 Ti GPU (12 GB). The model leverages CUDA 10.1 and cuDNN 8.1.1 for GPU acceleration. During training, a sequence length of 40 frames and a batch size of 8 are used. The Adam optimizer was employed with a learning rate of $1e^{-4}$. Additionally, weight decay was incorporated into the optimization process at the $50^{th}$ and $75^{th}$ epochs optimized over the categorical cross-entropy loss. Feature extraction followed the methodology outlined in~\cite{d2020transformer}. Input images are cropped to a size of $224 \times 224$ before being fed into the model for processing. Initially, NVGesture~\cite{molchanov2016online} and Briareo~\cite{manganaro2019hand} datasets provide the raw RGB, infrared, and depth images. We calculate the normals from depth maps and optical flow from the RGB images as described in~\cite{d2020transformer}. Experiments are performed separately for each input data and later fused using late fusion to get the multimodal input performance of the proposed model. Also, our model consists of 6 transformer stages.

\begin{table}[tb]
\caption{Comparitive results for multi-modalities on NVGestures dataset~\cite{molchanov2016online}. }
\centering
\begin{tabular}{M{2.76cm}M{3cm}M{1cm}}
\hline
 Method & Input modality & Accuracy \\ 
\hline
Two-st. CNNs~\cite{simonyan2014two} & Color + flow & 65.60\%\\
\hline
iDT ~\cite{vadisaction}& Color + flow & 73.00\% \\
\hline
R3D-CNN~\cite{molchanov2016online} & Color + flow & 79.30\%\\
R3D-CNN~\cite{molchanov2016online} & Color + depth + flow & 81.50\%\\
R3D-CNN~\cite{molchanov2016online} & Color + depth + ir & 82.00\%\\
R3D-CNN~\cite{molchanov2016online} & depth + flow & 82.40\%\\
R3D-CNN~\cite{molchanov2016online} & all & 83.80\%\\
\hline
MSD-2DCNN~\cite{fan2021multi}&Color+depth&84.00\% \\
\hline
8-MFFs-3f1c\cite{kopuklu2018motion}&Color + flow& 84.70\%\\
\hline
STSNN~\cite{zhang2020dynamic}&Color+flow& 85.13\%\\
\hline
PreRNN~\cite{yang2018making}& Color + depth&85.00\% \\
\hline

I3D~\cite{wang2016robust}& Color + depth &83.80\%\\
I3D~\cite{wang2016robust}& Color + flow &84.40\%\\
I3D~\cite{wang2016robust}& Color + depth + flow &85.70\%\\

\hline
GPM~\cite{fan2021multi}& Color + depth&86.10\% \\
\hline

MTUT\textsubscript{RGB-D}~\cite{abavisani2019improving}& Color + depth& 85.50\%\\
MTUT\textsubscript{RGB-D+flow}~\cite{abavisani2019improving}& Color + depth& 86.10\%\\
MTUT\textsubscript{RGB-D+flow}~\cite{abavisani2019improving}& Color + depth + flow& 86.90\%\\
\hline

Transformer~\cite{d2020transformer}& depth + normals &87.30\%\\
Transformer~\cite{d2020transformer}& Color + depth + normals + ir& 87.60\%\\
\hline

NAS2~\cite{yu2021searching}& Color + depth&86.93\% \\
NAS1+NAS2~\cite{yu2021searching} &Color + depth&88.38\% \\
\hline

GestFormer~\cite{garg2024gestformer}&depth + normals&82.78\%\\	
GestFormer~\cite{garg2024gestformer}&depth + color + ir\textbf{ }&84.24\%\\ 
\multirow{2}{*}{GestFormer~\cite{garg2024gestformer}}&depth + color + ir +&\multirow{2}{*}{85.62\%}\\ 
&normal&\\ 
\multirow{2}{*}{GestFormer~\cite{garg2024gestformer}}&depth + color + ir +&\multirow{2}{*}{85.85\%}\\
& normal + op&\\ 
\hline

\textbf{ConvMixFormer}&\textbf{depth + op}&\textbf{82.16\%}\\	
\textbf{ConvMixFormer}&\textbf{depth + normal+ ir }&\textbf{84.02\%}\\ 
\multirow{2}{*}{\textbf{ConvMixFormer}}&\textbf{depth + ir+ normal + flow }&\multirow{2}{*}{\textbf{85.49\%}}\\ 
\hline	
\end{tabular}
\label{tab3}
\end{table}

\subsection{Results and Discussion}
\begin{table}[tb]
\caption{Comparison of the results obtained for different modalities on Briareo dataset~\cite{manganaro2019hand}.}
\centering
\begin{tabular}{ccc}
\hline
 Method & Input modality & Accuracy   \\ 
\hline
C3D-HG~\cite{manganaro2019hand}& Color& 72.20\%\\
C3D-HG~\cite{manganaro2019hand}& depth& 76.00\%\\
C3D-HG~\cite{manganaro2019hand}& ir& 87.50\%\\

LSTM-HG~\cite{manganaro2019hand}&3D joint features &94.40\%\\
\hline
NUI-CNN~\cite{d2020multimodal}& depth + ir& 92.00\%\\
NUI-CNN~\cite{d2020multimodal}& Color + depth + ir& 90.90\%\\

\hline
Transformer~\cite{d2020transformer}& normals& 95.80\%\\
Transformer~\cite{d2020transformer}& depth + normals &96.20\%\\
Transformer~\cite{d2020transformer}&ir + normals &97.20\%\\

\hline	
GestFormer~\cite{garg2024gestformer} & ir&98.13\%\\
GestFormer~\cite{garg2024gestformer} & ir + normals &97.57\%\\

\hline
MVTN~\cite{garg2024mvtnmultiscalevideotransformer} & normals&98.26 \%\\
MVTN~\cite{garg2024mvtnmultiscalevideotransformer} & color + depth + normals &98.61\%\\
MVTN~\cite{garg2024mvtnmultiscalevideotransformer} &depth + ir + normals &98.61\%\\
\hline
\textbf{ConvMixFormer} & \textbf{Color}&\textbf{98.26} \%\\
\textbf{ConvMixFormer} & \textbf{Color + ir} &\textbf{98.61}\%\\
\textbf{ConvMixFormer} &\textbf{Color+ ir + normals} &\textbf{98.64}\%\\
\hline
\end{tabular}
\label{tab5}
\end{table}

\textbf{NVGesture:} 
The proposed ConvMixFormer model is evaluated on the NVGesture dataset for both single as well as multimodal inputs, keeping the same experimental set as in~\cite{d2020transformer}.  In Table~\ref{tab1}, the results for different modalities are compared with the conventional transformer~\cite{d2020transformer}. It can be seen from the table that ConvMixFormer achieves state-of-the-art results with a lesser number of parameters since the quadratic self-attention spatial token mixer is replaced with the conventional convolution layer to capture local features between different spatial tokens. From the table, we can infer that nearly the same accuracy is obtained when color images are given as input to the proposed model as compared to the transformer model in~\cite{d2020transformer}. With a drop of 1-2.5\%, the proposed model shows comparable results for other inputs (normal, depth, IR, optical flow) with half a number of parameters. Among all the inputs, the best accuracy for single modal data is obtained for depth images with 80.83\% accuracy. Nearly the same accuracy is obtained from the normals since they are derived from depth images.  

From the table, we also conclude that the performance of the proposed model increases with the increase in the number of inputs which is obtained using late fusion of the probability scores. For two inputs, our model achieves the best accuracy when depth map output probabilities are fused with an optical flow output probability score which is 82.16\%. From this, we can infer that adding optical flow to depth images increases the accuracy by nearly 2\% when compared with the single depth map input. Similarly, when the scores obtained from infrared and normal images are fused with the depth maps, 84.02\% accuracy is obtained which is again nearly a 2\% rise in accuracy from the 2 input modalities. Further, a rise in nearly 1.5\% accuracy is witnessed when optical flow is fused with the best accuracy combination of the 3 modality inputs giving 85.49\% accuracy. From the experiments until 4 modality inputs, we can conclude that increasing the number of inputs increases the accuracy from 80.83\% to 85.49\%. However, further increasing the input modalities to 5 doesn't improve the performance of ConvMixFormer.

Additionally, we have also compared the performance of our proposed ConvMixFormer with the state-of-the-art methods on a single modality in Table~\ref{tab2}. After careful analysis, we conclude from the table that ConvMixFormer achieves better performance than GestFormer~\cite{garg2024gestformer} which is a transformer-based approach with an even lesser number of parameters as shown in Table~\ref{tab24} and comparable results with conventional Transformer~\cite{d2020transformer} and other methods.  Similarly, we have also compared the results for multimodal inputs in Table~\ref{tab3}. 

\textbf{Briareo:} 
Similar to the NVGesture dataset, we have also compared the performance of the proposed ConvMixFormer on the Briareo dataset for single and multimodal inputs with the conventional transformer~\cite{d2020transformer} as shown in Table~\ref{tab1}.  We can infer from the table that ConvMixFormer can efficiently mix tokens for the Briareo dataset achieving the best performance on RGB images with an accuracy of 98.26\%. It can also be seen in the table that for each set of experiments, ConvMixFormer achieves better performance than Transformer~\cite{d2020transformer} for the Briareo dataset. An increment of 8.45\%, 5.22\%, 2.23\%, and 1.85\% is evident from color, depth, infrared, and normal inputs respectively, when given as single mode input. For two inputs, the best accuracy of 98.61\% is obtained when RGB images are fused with infrared images. Further, increasing the number of inputs to 3 increases the accuracy to 98.64\%, when normals are fused with the best accuracy combination for 2 inputs (RGB, IR). However, the addition of inputs to 4 and 5 modalities does not show an improvement in the performance. We have also noticed in the table that for each modality, the overall performance is better. 

\begin{table}[tb]
\caption{Ablation study on the proposed ConvMixFormer model.}
\centering
\begin{tabular}{M{1cm}|M{3.7cm}|M{1.15cm}|M{1cm}}
\hline
Baseline & Module & Accuracy & Params (M)\\  \hline
BL1&ConvMix with FFN & 77.50\%& 17.51\\
BL2&ConvMix with GDFN  & 81.25\% & 23.83\\
\multirow{2}{*}{BL3}&ConvMix with GDFN less & \multirow{2}{*}{\textbf{80.83\%} }& \multirow{2}{*}{\textbf{13.57}}\\
&   (ConvMixFormer) & &\\
\hline
\end{tabular}
\label{tab25}
\end{table}
Further, we also compared the results obtained by ConvMixFormer with other methods on different modalities in Table~\ref{tab5}. It can be noted from the table that ConvMixFormer achieves benchmark results compared to other methods with an accuracy of 98.26\%. ConvMixFormer with single modal input can even outperform other methods with multimodal inputs. Further, increment in modality gives rise in the accuracy to 98.64\%.

\subsection{Ablation Study}
The proposed ConvMixFormer has 3 baselines (BL1, BL2, and BL3) as shown in Table~\ref{tab25}.  Baseline BL1 is the convolution mixer-based transformer model with a feed-forward network (FFN)  same as in~\cite{d2020transformer}. Baseline BL2 explores the convolution mixer with the use of a gating mechanism in the feed-forward network. In BL2, the gating network in FFN is implemented similar to~\cite{khan2024spectroformer}. This involves projecting the input channels to twice the number of hidden channels which are then re-projected to the same number of channels as the input channel.  Baseline BL3 is the same as the BL2. The only difference is that, in the gating network,  the input channels are now projected to the same number of parameters as the hidden channels. Since it involves fewer parameters compared to BL2, we have named it ConvMix with GDFN less which is our proposed ConvMixFormer model. 

An initial experiment that shows the performance of ConvMix with FFN  obtains  77.50\% accuracy with 17.51M parameters. To further enhance the model performance, we have experimented with the use of a gating mechanism in the FFN which increases the performance of the model 81.25\%. Though a significant rise in accuracy is seen in BL2 from BL1, at the same time, the number of parameters also increases drastically to 23.83M. The rise in accuracy was not significant compared to the rise in parameters. So, we performed another experiment with a smaller number of parameters in the Gated Depthwise Feed Forward Network i.e. BL3. BL3 achieves lesser accuracy than BL2 but it is still improved as compared to BL1 with an even lesser number of parameters than BL1 and BL2 both. So, we choose BL3 as our proposed ConvMixFormer model with 13.57M parameters and 80.83\% accuracy. Thus, Table~\ref{tab25} validates our design considerations.

\subsection{Parameter Efficiency}
We have also compared the ConvMixFormer model with other state-of-the-art methods in terms of the number of parameters and MACs (Multiply-Accumulate Operations). Table~\ref{tab24} shows that the proposed ConvMixFormer significantly reduces the number of parameters compared to other methods. The parameters reduce to nearly half as compared to the transformer model~\cite{d2020transformer} and GestFormer~\cite{garg2024gestformer}, which are also transformer-based models for gesture recognition tasks. Despite the reduction in parameters, ConvMixFormer maintains comparable performance compared to transformer models for the NVGesture dataset and improved performance for the Briareo dataset. MACs are also comparatively less than other methods, which further highlights its efficiency in terms of computational resources.

\begin{table}[tb]
\caption{Comparison in terms of the number of parameters (M)) and MACs (G).  MACs are counted by fvcore library. }
\centering
\begin{tabular}{c|cc}
\hline
Methods&Params (M)& MACs (G)\\  \hline

NAS1~\cite{yu2021searching} &93.90&60.44 \\
NAS2~\cite{yu2021searching} &251.40& 116.20\\
ResNeXt-101~\cite{kopuklu2019real} &52.28&- \\
R3D-CNN~\cite{molchanov2016online} & 38.00&-\\
NUI-CNN~\cite{d2020multimodal}& 28.00& -\\
C3D-HG~\cite{manganaro2019hand} & 26.70&-\\
Transformer\cite{d2020transformer} & 24.30&62.92\\
GestFormer~\cite{garg2024gestformer}&24.08&60.40\\
MVTN~\cite{garg2024mvtnmultiscalevideotransformer}&19.55& 60.22\\
\textbf{ConvMixFormer}&\textbf{13.57}&\textbf{59.98}\\
\hline
\end{tabular}
\label{tab24}
\end{table}	

\section{Conclusion}
In this paper, we propose a novel and resource-efficient ConvMixFormer for dynamic hand gesture recognition, which includes a novel convolution token mixer for efficiently extracting spatial features from the input features. To further enhance the model's ability to learn more powerful representations, we employ Gated Deconv FFN. By replacing the self-attention module with a convolution layer, a significant decrease in the number of parameters can be observed. The proposed ConvMixFormer achieves comparable results for the NVGesture dataset and outperforms traditional transformer and other methods for the Briareo dataset with nearly half the number of parameters and less complexity.   Finally, we can conclude that convolution-based token mixers are well-suited for gesture recognition tasks.

{\small
\bibliographystyle{ieee_fullname}
\bibliography{egbib}
}

\end{document}